# Graph-incorporated Latent Factor Analysis for High-dimensional and Sparse Matrices

Di Wu[1,2], Yi He[3], and Xin Luo[4]

1 Chongqing Institute of Green and Intelligent Technology, Chinese Academy of Sciences, Chongqing 400714, China.
2 Institute of Artificial Intelligence and Blockchain, Guangzhou University, Guangzhou 510006, Guangdong, China.
3 Old Dominion University, Norfolk, Virginia 23462, USA.
4 School of Computer Science and Technology, Chongqing University of Posts and Telecommunications, Chongqing 400065, China.
(e-mail: wudi.cigit@gmail.com)

**Abstract.** A High-dimensional and sparse (HiDS) matrix is frequently encountered in a big data-related application like an e-commerce system or a social network services system. To perform highly accurate representation learning on it is of great significance owing to the great desire of extracting latent knowledge and patterns from it. Latent factor analysis (LFA), which represents an HiDS matrix by learning the low-rank embeddings based on its observed entries only, is one of the most effective and efficient approaches to this issue. However, most existing LFA-based models perform such embeddings on a HiDS matrix directly without exploiting its hidden graph structures, thereby resulting in accuracy loss. To address this issue, this paper proposes a graph-incorporated latent factor analysis (GLFA) model. It adopts two-fold ideas: 1) a graph is constructed for identifying the hidden high-order interaction (HOI) among nodes described by an HiDS matrix, and 2) a recurrent LFA structure is carefully designed with the incorporation of HOI, thereby improving the representation learning ability of a resultant model. Experimental results on three real-world datasets demonstrate that GLFA outperforms six state-of-the-art models in predicting the missing data of an HiDS matrix, which evidently supports its strong representation learning ability to HiDS data.

**Keywords:** High-dimensional and Sparse Matrix, Latent Factor Analysis, Missing Data Prediction, Graph Structure, Graph Representation Learning.

## 1 Introduction

In industrial applications, matrices are widely adopted to describe the relationships between two types of entities. While in many big-data-related applications like bioinformatic networks[1], e-commerce systems [2], and wireless sensor networks[3], since the relationships among numerous entities are unlikely to be fully observed in practice [4, 5], matrices from these applications are commonly high-dimensional and sparse (HiDS). Despite these HiDS matrices are usually extremely sparse, they contain rich latent knowledge and patterns [6], e.g., users' potential preferences on items in electronic commerce systems. Hence, how to precisely extract valuable knowledge and patterns from such HiDS matrices becomes a hot yet thorny issue in industrial applications [7-15].

Recently, latent factor analysis (LFA) [16-21] has become one of the most popular and successful approaches to this issue. Given an HiDS matrix, an LFA model represents it by learning two low-rank embedding matrices based on its observed entries only [8, 22-26]. However, most existing LFA-based models learn the embeddings from an HiDS matrix directly without exploiting its hidden graph structures [27, 28], thereby the learned embeddings may not be sufficient to accurately represent it.

Recently, Zhou *et al.* propose the *Deep Forest* to alleviate some drawbacks of deep neural networks (DNNs) [29]. Its principle is to gradually enhance a basic model by enriching its inputs via a carefully designed multilayer learning structure, where each layer is a basic model and its additional inputs come from the prediction outputs of its preceding basic model. From this point of view, it becomes possible to gradually enhance an LFA model by recurrently enriching its inputs via a carefully designed recurrent (or multilayer) learning structure, where the additional inputs come from the prediction outputs of an LFA model. In particular, if we can recurrently incorporate the aforementioned HOI information into the prediction outputs of an LFA model, we may enhance an LFA model's representation learning ability to an HiDS matrix.

To this end, this paper proposes a graph-incorporated latent factor (GLFA) model. It consists of two main parts. One part is to construct a graph from a given HiDS matrix for identifying its hidden HOI among nodes. The other one is to design a recurrent LFA structure to gradually enhance an LFA model by incorporating the identified HOI information into the embedding learning process. Experimental results on three real-world datasets verify that the proposed GLFA model outperforms six state-of-the-art models in representing HiDS data, including four LFA-based and two DNNs-based models.

The main contributions of this work include:

- A GLFA model that can accurately represent an HiDS matrix to predict its missing data is proposed;
- Algorithm design and analysis are given for the proposed GLFA model;
- Extensive experiments on three real-world datasets are conducted to evaluate the proposed GLFA model, including comparing it with six state-of-the-art models and analyzing its characteristics.

## 2    Related Work

After LFA achieves great success in the high-profile competition of Netflix Prize [16], the LFA-based models for HiDS data analysis are springing up in the past decade. So far, various improved LFA-based models are proposed, including a non-negative constraint one [5], a dual-regularization-based one [30], a generalized momentum-based one [10], a probabilistic one [31], a neighborhood-and-location integrated one [32], a joint recommendation one [33], a confidence-driven one [34], a content features-based one[35], etc. In particular, although some other types of approach, like the DNNs-based one, show some promising advances in processing HiDS data related applications such as recommender systems, S. Rendle *et al.* demonstrate that the LFA model still has competitive performance [36]. However, given an HiDS matrix, the existing LFA-based models ignore its hidden graph structures that contains rich latent knowledge and patterns [27, 28] during their embedding learning process. Different from these models, the proposed GLFA model can incorporate the HOI information of a graph constructed from an HiDS matrix into its embedding learning process via a recurrent LFA structure, thereby achieving the accurate representation of an HiDS matrix.

Recently, DNNs attract much attention in processing HiDS data as their powerful representation learning ability [37, 38], especially in the field of recommender systems. Zhang *et al.* conduct a detailed review regarding this issue [39]. Up to now, many related DNNs-based models are proposed. Representative approaches include autoencoder-based one [40], a variational autoencoder-based one [41], a hybrid autoencoder-based one [42], a neural collaborative filtering-based one [37], a neural factorization-based one [43], etc. Besides, regarding exploiting the hidden graph structures of an HiDS matrix, Berg et al. propose a graph autoencoder-based model [27] and Wang *et al.* propose a neural graph-based model [44]. However, the graph autoencoder-based model requires not only the HiDS matrix but also additional features of entities, and the neural graph-based model ignores the weights (values of entries of the HiDS matrix) of the graph structures. While the proposed GLFA model fully exploits the hidden graph structures without the requirement of any additional information. Besides, when representing an HiDS matrix, a DNNs-based model's performance is achieved at the cost of high computation burden [29]. In comparison, the proposed GLFA model has a higher computational efficiency because its basic component is the highly efficient LFA model [36] trained only on observed entries of an HiDS matrix.

Note that in Section 5.2, some DNNs-based models mentioned above are not compared because they are defined on different data sources, such as models proposed in [24, 28, 30-31] only focus on binary coded HiDS matrix and ones proposed in [12, 25] rely on additional review information.

## 3    Preliminaries

***Definition* 1** (*HiDS Matrix*). Given two types of entity sets $U$ and $I$, $R$ is a $|U|\times|I|$ matrix where each entry $r_{u,i}$ denotes the relationship between entity $u$ ($u \in U$) and entity $i$ ($i \in I$). $\Psi$ and $\Gamma$ respectively denote the sets of observed and unobserved entries of $R$. $R$ is a high-dimensional and sparse (HiDS) matrix with $|\Psi| \ll |\Gamma|$.

***Definition* 2** (*LFA Model*). Given $R$, a latent factor analysis (LFA) model is to learn two embedding matrices $X^{|U|\times f}$ and $Y^{|I|\times f}$ to make $R$'s rank-$f$ approximation $\hat{R}$ by minimizing the distances between $R$ and $\hat{R}$ on $\Psi$, where $f$ is the embedding dimension with $f \ll \min\{|U|, |I|\}$ and $\hat{R}$ can be obtained by $\hat{R}=XY^{\mathrm{T}}$. $\hat{R}$'s each entry $\hat{r}_{u,i}$ is the corresponding prediction for $r_{u,i}$.

From definition 2, we see that an LFA model works by designing an objective function to minimize the distances between $R$ and $\hat{R}$ on $\Psi$. Commonly, Euclidean is adopted as the metric distance to design such an objective function as follows:

$$\arg\min_{X,Y} \varepsilon(X, Y) = \frac{1}{2}\left\|\Omega \odot (R - \hat{R})\right\|_F^2 = \frac{1}{2}\left\|\Omega \odot (R - XY^T)\right\|_F^2, \quad (1)$$

where $\|\cdot\|_F$ denotes the Frobenius norm of a matrix, $\odot$ denotes the Hadamard product (the component-wise multiplication), and $\Omega$ is a $|U|\times|I|$ binary index matrix given by:

$$\Omega_{u,i} = \begin{cases} 1 & \text{if } r_{u,i} \text{ is observed} \\ 0 & \text{otherwise} \end{cases}, \quad (2)$$

where $\Omega_{u,i}$ denotes the $u$-th row and $i$-th column entry of $\Omega$. Solving (1) is an ill-posed problem because of overfitting. To address this, Tikhonov regularization is widely employed to incorporate into (1) as follows [16]:

$$\arg\min_{X,Y} \varepsilon(X, Y) = \frac{1}{2}\left\|\Omega \odot (R - XY^T)\right\|_F^2 + \frac{\lambda}{2}\left(\|X\|_F^2 + \|Y\|_F^2\right), \quad (3)$$

where $\lambda$ is the regularization hyper-parameter. In general, (3) can be minimized by an optimization algorithm like stochastic gradient descent (SGD) [45, 46].

## 4  Proposed GLFA Model

Following the example shown in Fig. 1 and the principle of Deep Forest [29], we propose the GLFA model, as shown in Fig. 1. It consists of two main parts: one is to construct a weighted interaction graph from a given HiDS matrix for identifying its hidden high-confidence HOI, and the other one is to design a recurrent LFA structure to incorporate the identified high-confidence HOI information into the embedding learning process.

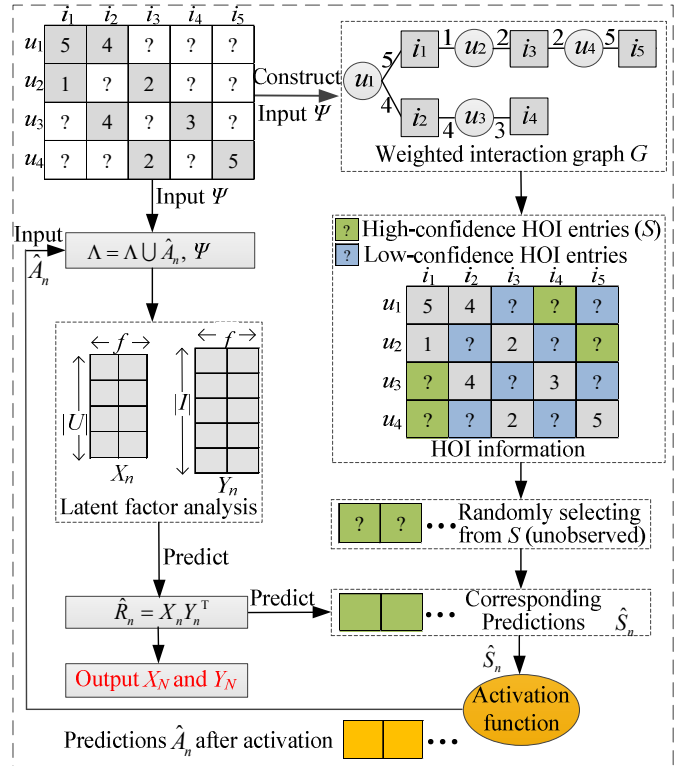

**Fig. 1.** The proposed GLFA model.

### 4.1  Working flow of GLFA

GLFA works as follows:
1) Input: inputting $\Psi$ of $R$ as the initial inputs;
2) Part 1: constructing weighted interaction graph $G$ based on $\Psi$ (definition 3);
3) Part 1: identifying all the high-confidence HOIs based on $G$ (definitions 4 and 5);

4) Part 1: putting all the identified high-confidence HOIs into set $S$ (definition 6);
5) Part 2 at $n$-th training: training embedding matrices $X_n$ and $Y_n$ based on $\Psi$ and $\Lambda$, where $n \in \{1, \dots, N\}$, $N$ is presetting maximum times of recurrent training and $\Lambda$ contains all the predictions for the elements in $S$ by the previous $n$-1 times of recurrent training. Note that $\Lambda$ is empty when $n$=1;
6) Part 2 at $n$-th training: randomly selecting elements from $S$ to generate a set $S_n$, $S_n \subseteq S$, predicting all the elements of $S_n$ via $\hat{R}_n = X_n Y_n^T$, and putting the predictions into set $\hat{S}_n$, $S \leftarrow S-S_n$;
7) Part 2 at $n$-th training: inputting $\hat{S}_n$ into a nonlinear activation function, its output is $\hat{A}_n$ set;
8) Part 2 at $n$-th training: putting $\hat{A}_n$ into $\Lambda$ as the input for the next training;
9) Part 2 with recurrent training: repeating steps 5)–8) until to reach $N$;
10) Output: outputting the learned $X_N$ and $Y_N$.

## 4.2 Constructing a Graph from an HiDS Matrix to Identify HOI among nodes.

We first give some related definitions regarding the proposed GLFA model.

***Definition 3*** (*Weighted Interaction Graph of R*). Given $R$, let $G=(U, I, E, R)$ be $R$'s weighted interaction graph, where $U$ and $I$ are seen as two types of vertex sets, $E$ is the edge set, and $R$ is seen as the weight matrix of edges. Each edge $e_{u,i}$ of $E$ denotes that a vertex $u$ ($u \in U$) connects to a vertex $i$ ($i \in I$). Each entry $r_{u,i}$ of $R$ denotes the weight of an edge $e_{u,i}$. Note that $E$ can be easily obtained from $R$ and $r_{u,i}$ is unknown denotes that there is no edge between $u$ and $i$.

***Definition 4*** (*HOI of G*). Given $G$, one indirect interaction $(u, i)$ is $G$'s high-order interaction (HOI) if the shortest path from $u$ to $i$ pass through at least $p$-1 ($p \geq 2$) vertexes from $U$. $p$ denotes the $p$-th HOI.

***Definition 5*** (*High/low-confidence HOI*). Given $G$, let $(u, i)$ be a $p$th HOI. Supposing there are several different paths from $u$ to $i$. Among each of the several different paths, if there is any vertex from $I$ is connected with two different weight edges to vertexes from $U$, $(u, i)$ is low-confidence; otherwise, it is high-confidence.

***Definition 6*** (*High-confidence HOI set*). Given $G$, the high-confidence HOI set $S$ consists of all the high-confidence HOIs among $G$. Note that each element of $S$ is directly unobserved by $R$.

Let's illustrate by an example inserted in Fig. 2. Given $R$ with four users and five items (top left corner), its $G$ is constructed as the top right corner. Based on $G$, we can identify high/low-confidence HOI. For example, $u_1$'s HOI items are $i_3$ (2-nd), $i_4$ (2-nd), and $i_5$ (3-nd), where $(u_1, i_4)$ is high-confidence because the weight between $u_3$ (connected with $i_4$) and $i_2$ is 4 that is the same as that between $u_1$ and $i_2$, $(u_1, i_3)$ and $(u_1, i_5)$ are low-confidence because from $i_3$ and $i_5$ to $u_1$, $i_1$ is connected with two different weight edges (weight 5 and weight 1). Finally, the high-confidence HOI set of $G$ is $S=\{(u_1, i_4), (u_2, i_5), (u_3, i_1), (u_4, i_1)\}$.

## 4.3 Recurrently Training GLFA with SGD

To explain how to train GLFA, we give the $n$-th recurrent training process as a general description. To efficiently train GLFA, we first extend (3) into the single latent-factor-dependent form as follows:

$$\arg\min_{X_n, Y_n} \varepsilon(X_n, Y_n) = \frac{1}{2} \sum_{r_{u,i} \in \Psi} \left( r_{u,i} - \sum_{\tau=1}^{f} x_{u,\tau}^n y_{i,\tau}^n \right)^2 + \frac{\lambda}{2}\left(\|X_n\|_F^2 + \|Y_n\|_F^2\right), \quad (4)$$

where $X_n$ and $Y_n$ are the embedding matrices at $n$-th recurrent training, $x_{u,\tau}^n$ is $X_n$'s $u$-th row and $\tau$-th column entry, and $y_{i,\tau}^n$ is $Y_n$'s $i$-th row and $\tau$-th column entry, respectively. To incorporate the identified high-confidence HOI information into an LFA model, we transform (4) as follows.

$$\arg\min_{X_n, Y_n} \varepsilon(X_n, Y_n) = \underbrace{\frac{1}{2} \sum_{r_{u,i} \in \Psi} \left( r_{u,i} - \sum_{\tau=1}^{f} x_{u,\tau}^n y_{i,\tau}^n \right)^2}_{\text{Training on input HiDS matrix}} + \underbrace{\frac{1}{2}\alpha \sum_{\bar{r}_{u,i} \in \Lambda} \left( \bar{r}_{u,i} - \sum_{\tau=1}^{f} x_{u,\tau}^n y_{i,\tau}^n \right)^2}_{\text{Training on high-confidence HOIs}} + \frac{\lambda}{2}\left(\|X_n\|_F^2 + \|Y_n\|_F^2\right), \quad (5)$$

where $\alpha$ is the aggregation coefficient, $\Lambda$ is a set that contains all the predictions for the high-confidence HOIs by the previous $n$-1 times of recurrent training, and $\bar{r}_{u,i}$ is an element of $\Lambda$ corresponding to $r_{u,i}$. Note that when $n$=1, $\Lambda$ is empty. Then, we consider the instant of (5) on a single element $\forall \bar{r}_{u,i} \in \Lambda$ or $\forall r_{u,i} \in \Psi$ as follows:

$$\begin{cases} \text{if on } r_{u,i}: \varepsilon_{u,i}^n = \dfrac{1}{2}\left(r_{u,i} - \sum_{\tau=1}^{f} x_{u,\tau}^n y_{i,\tau}^n\right)^2 + \dfrac{\lambda}{2}\left(\sum_{\tau=1}^{f}(x_{u,\tau}^n)^2 + \sum_{\tau=1}^{f}(y_{i,\tau}^n)^2\right) \\ \text{if on } \bar{r}_{u,i}: \varepsilon_{u,i}^n = \dfrac{1}{2}\alpha\left(\bar{r}_{u,i} - \sum_{\tau=1}^{f} x_{u,\tau}^n y_{i,\tau}^n\right)^2 + \dfrac{\lambda}{2}\left(\sum_{\tau=1}^{f}(x_{u,\tau}^n)^2 + \sum_{\tau=1}^{f}(y_{i,\tau}^n)^2\right) \end{cases}. \quad (6)$$

Next, we adopt SGD to optimize (6) with respect to each single embedding element:

$$\forall \tau \in \{1,2,...,f\}, \text{on } r_{u,i} \text{ or } \bar{r}_{u,i}: \begin{cases} x_{u,\tau}^n \leftarrow x_{u,\tau}^n - \eta \dfrac{\partial \varepsilon_{u,i}^n}{\partial x_{u,\tau}^n} \\ y_{i,\tau}^n \leftarrow y_{i,\tau}^n - \eta \dfrac{\partial \varepsilon_{u,i}^n}{y_{i,\tau}^n} \end{cases}, \quad (7)$$

where $\eta$ is the learning rate. By extending (7), we have the training rules as follows:

$$\forall \tau \in \{1,2,...,f\}: \begin{cases} \text{if on } r_{u,i}, \begin{cases} x_{u,\tau}^n \leftarrow x_{u,\tau}^n + \eta y_{i,\tau}^n\left(r_{u,i} - \sum_{\tau=1}^{f} x_{u,\tau}^n y_{i,\tau}^n\right) - \eta\lambda x_{u,\tau}^n \\ y_{i,\tau}^n \leftarrow y_{i,\tau}^n + \eta x_{u,\tau}^n\left(r_{u,i} - \sum_{\tau=1}^{f} x_{u,\tau}^n y_{i,\tau}^n\right) - \eta\lambda y_{i,\tau}^n \end{cases} \\ \text{if on } \bar{r}_{u,i}, \begin{cases} x_{u,\tau}^n \leftarrow x_{u,\tau}^n + \eta y_{i,\tau}^n\alpha\left(\bar{r}_{u,i} - \sum_{\tau=1}^{f} x_{u,\tau}^n y_{i,\tau}^n\right) - \eta\lambda x_{u,\tau}^n \\ y_{i,\tau}^n \leftarrow y_{i,\tau}^n + \eta x_{u,\tau}^n\alpha\left(\bar{r}_{u,i} - \sum_{\tau=1}^{f} x_{u,\tau}^n y_{i,\tau}^n\right) - \eta\lambda y_{i,\tau}^n \end{cases} \end{cases}. \quad (8)$$

After each element in $\Psi$ and $\Lambda$ is trained with (8), $X_n$ and $Y_n$ are learned. Then, we employ them to predict the elements of $S_n$ that are randomly selected from $S$, $S_n \subseteq S$, via $\hat{R}_n = X_n Y_n^T$, where $S \leftarrow S-S_n$ and $\hat{r}_{u,i}^n$ is the entry of $\hat{R}_n$ at $u$-th row and $i$-th column. Let $\hat{S}_n$ be a set whose elements are the predictions for the elements of $S_n$. $\hat{S}_n$ is input into a nonlinear activation function as follows.

$$\forall \hat{r}_{u,i}^n \in \hat{S}_n : \tilde{r}_{u,i}^n = \begin{cases} r_{min} + 1/\left(1 + e^{-\hat{r}_{u,i}^n}\right) & \text{if } \hat{r}_{u,i}^n < r_{min} \\ r_{max} / \left(1 + e^{-\hat{r}_{u,i}^n}\right) & \text{if } \hat{r}_{u,i}^n > r_{max} \\ \hat{r}_{u,i}^n & \text{otherswise} \end{cases}, \quad (9)$$

where $r_{min}$ and $r_{max}$ denote the minimum and maximum value, respectively. Obviously, $\hat{r}_{u,i}^n < r_{min}$ or $\hat{r}_{u,i}^n > r_{max}$ is not correct, (9) has the function of resetting the extremely unreasonable predictions. Let $\hat{A}_n$ denote all the outputs of (9), i.e., $\hat{A}_n$ is composed of $\tilde{r}_{u,i}^n$. Then, we put $\hat{A}_n$ into $\Lambda$ as the input for the next training, i.e.,

$$\Lambda = \Lambda \cup \hat{A}_n. \quad (10)$$

Finally, after $N$ times of recurrent training, output the learned embedding matrices $X_N$ and $Y_N$, which can be employed to make $R$'s rank-$f$ approximation $\hat{R}$ by $\hat{R} = X_N Y_N^T$.

## 5    Experiments and Results

In the experiments, we aim at answering the following research question: Does the proposed GLFA model outperform state-of-the-art LFA-based and DNNs-based models in representing an HiDS matrix?

### 5.1    General Settings

**Datasets.** Three benchmark real-world HiDS datasets are adopted in the experiments. Their properties are summarized in Table 1. Ml1m is collected by MovieLens[1] system that is an online movie platform. Yahoo is collected by

---
[1] https://grouplens.org/datasets/movielens/

Yahoo!Music[2] and records a music community's preferences for various songs. Goodbook[3] is collected by a real-world online book site with six million ratings for ten thousand most popular books. For each dataset, we randomly select its 20% know data as the training set and the remaining 80% ones are treated as the testing set.

Table 1. Properties of all the HiDS datasets.

| Name | $|U|$ | $|I|$ | $|\Psi|$ | Density* |
|---|---|---|---|---|
| Ml1m | 6040 | 3706 | 1,000,209 | 4.47% |
| Yahoo | 15400 | 1000 | 365,704 | 2.37% |
| Goodbooks | 53424 | 10000 | 5,976,480 | 1.12% |

*Density denotes the percentage of known data in the HiDS matrix.

**Evaluation Protocol.** In many big-data-related applications, it is highly significant to recover the unknown relationships among an HiDS matrix. Hence, the task of missing data prediction is widely adopted as the evaluation protocol of representing an HiDS matrix [47-53]. To evaluate the prediction accuracy, root mean squared error (RMSE) and mean absolute error (MAE) are usually adopted as follows [10]:

$$RMSE = \sqrt{\left(\sum_{(w,j)\in \Upsilon}(r_{w,j} - \hat{r}_{w,j})^2\right)\Big/|\Upsilon|}, \quad MAE = \left(\sum_{(w,j)\in \Upsilon}|r_{w,j} - \hat{r}_{w,j}|_{abs}\right)\Big/|\Upsilon|,$$

where $|\cdot|_{abs}$ indicate the absolute value and $\Upsilon$ indicates the testing set.

**Baselines.** We compare GLFA with six state-of-the-art baselines. They are briefly described in Table 2, where four baselines are LFA-based model (BLF, DLF, NLF, and FNLF) and two ones (AutoRec and DCCR) are DNNs-based model.

Table 2. The descriptions of state-of-the-art comparison baselines.

| Model | Description |
|---|---|
| BLF | A basic LFA model (2009) [16]. It wins the Netflix Prize[4] and has been extensively applied to represent HiDS data. |
| DLF | A deep LFA model (2019) [14]. It has a deep structure to play a role like regularization for improving BLF's representation learning ability. |
| NLF | A non-negative LFA model (2014). It introduces the non-negative constraints into an LFA mode for efficiently representing non-negative HiDS data [5]. |
| FNLF | A fast non-negative LFA model (2021) [10]. It improves NLF's representation learning ability by considering the momentum effects. |
| AutoRec | An autoencoder-based framework (2015) [40]. It is the representative model of DNNs-based model to represent HiDS data from recommender systems. |
| DCCR | A DNNs-based model (2019) [42]. It improves AutoRec's representation learning ability by using two different neural networks. |

## 5.2 Performance Comparison

**Comparison of RMSE and MAE.** Table 3 presents the comparison results. To better understand these comparison results, we also conduct the statistical analysis of the win/loss counts that GLFA versus comparison models. The win/loss counts are summarized in the last row of Table 3. We observe that GLFA wins the four LFA-based models in most cases (loses only one case). When compared with the two DNNs-based models, GLFA evidently wins them. Therefore, these observations show that GLFA has lower RMSE/MAE than the six comparison models.

Moreover, to check whether GLFA has significantly lower RMSE/MAE than each single model, we conduct the Wilcoxon signed-ranks test on the comparison results of Table 3. Wilcoxon signed-ranks test has three indicators—$R+$, $R-$, and $p$-value [54]. The larger $R+$ value indicates a lower RMSE/MAE and the $p$-value indicates the signifi-cance level. Table 4 records the tested results, where we see that all the hypotheses are accepted, which verifies that GLFA has a significantly higher prediction accuracy than the six comparison models.

---

[2] https://webscope.sandbox.yahoo.com/catalog.php?datatype=r
[3] http://fastml.com/goodbooks-10k-a-new-dataset-for-book-recommendations/
[4] https://www.netflixprize.com/

**Table 3.** The comparison results on RMSE/MAE including win/loss counts, where • indicates GLFA has a lower RMSE/MAE than the comparison models.

| Dataset | Metric* | BLF | DLF | NLF | FNLF | AutoRec | DCCR | GLFA |
|---|---|---|---|---|---|---|---|---|
| Ml1m | RMSE | 0.9175• | 0.9163• | 0.9290• | 0.9282• | 0.9169• | 0.9151• | 0.9126 |
|  | MAE | 0.7243• | 0.7237• | 0.7336• | 0.7280• | 0.7271• | 0.7253• | 0.7205 |
| Yahoo | RMSE | 1.4371• | 1.4244• | 1.4302• | 1.4313• | 1.4428• | 1.4382• | 1.4188 |
|  | MAE | 1.0844• | 1.0849• | 1.0868• | 1.0831• | 1.0925• | 1.0899• | 1.0794 |
| Goodbooks | RMSE | 0.8672 | 0.8688• | 0.8711• | 0.8701• | 0.8790• | 0.8758• | 0.8674 |
|  | MAE | 0.6812• | 0.6797 | 0.6778 | 0.6784 | 0.6802• | 0.6824• | 0.6800 |
| Win/Loss |  | 5/1 | 5/1 | 5/1 | 5/1 | 6/0 | 6/0 | — |

\* A **lower** RMSE/MAE value indicates a higher prediction accuracy.

**Table 4.** Results of Wilcoxon signed-ranks test on RMSE/MAE of Table 3.

| Comparison | $R+$ | $R-$ | $p$-value* |
|---|---|---|---|
| GLFA vs. BLF | 20 | 1 | **0.0313** |
| GLFA vs. DLF | 20 | 1 | **0.0313** |
| GLFA vs. NLF | 20 | 1 | **0.0313** |
| GLFA vs. FNLF | 20 | 1 | **0.0313** |
| GLFA vs. AutoRec | 21 | 0 | **0.0156** |
| GLFA vs. DCCR | 21 | 0 | **0.0156** |

\* The accepted hypotheses with a significance level of 0.05 are highlighted.

**Comparison of CPU Running Time.** Fig. 2 records the CPU running time of each model tested on all the datasets. First, we see that GLFA costs more CPU running time than the four LFA-based models. One reason is that GLFA trains some extra data due to its recurrent LFA structure. However, we see that when compared with the two DNNs-based models, GLFA costs much less CPU running time. The main reason is that GLFA only takes the observed entries of HiDS matrix as input while AutoRec and DCCR take the complete data to do that.

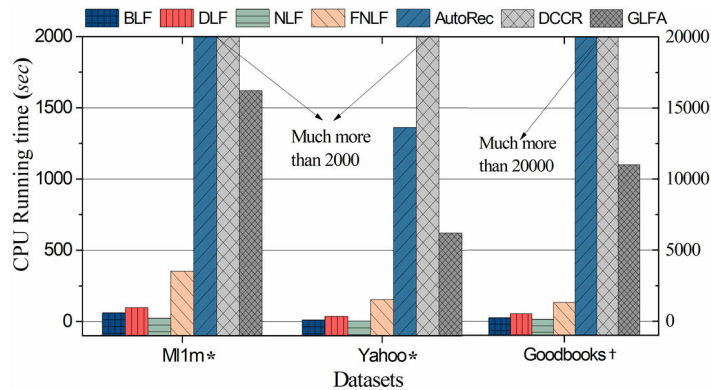

**Fig. 2.** The comparison of CPU running time, where * denotes that Ml1M and Yahoo are left tick label, and † denotes that Goodbooks is right tick label.

## 6  Conclusion

This paper proposes a graph-incorporated latent factor analysis (GLFA) model for accurately representing an HiDS matrix. By recurrently incorporating high-order interaction (HOI) information among nodes, which is identified from a weighted interaction graph of the HiDS matrix, into GLFA's embedding learning process, its representation learning ability is enhanced. In the experiments, GLFA is compared with six state-of-the-art models on three real-world datasets, the results demonstrate that it outperforms all the comparison models in predicting the missing data of an HiDS matrix. Moreover, GLFA achieves a much higher computational efficiency than the two DNNs-based comparison models. In the future, we plan to use GLFA for more machine learning tasks[55, 56].